\documentclass[runningheads]{llncs}

\usepackage{iciap}

\usepackage{iciapabbrv}

\usepackage{graphicx}
\usepackage{booktabs, microtype}

\usepackage[accsupp]{axessibility}  

\usepackage{hyperref}

\usepackage{orcidlink}

\usepackage{pifont}
\usepackage{wrapfig}
\usepackage{makecell}
\usepackage{multicol}
\usepackage[font=small,labelfont=bf,tableposition=top]{caption}
\DeclareCaptionLabelFormat{andtable}{#1~#2  \&  \tablename~\thetable}

\def\metricfull{Localized VQAScore\xspace}
\def\metricshort{L-VQAScore\xspace}

\usepackage[most]{tcolorbox}

\begin{document}

\title{Evaluating Attribute Confusion in Fashion Text-to-Image Generation}


\author{Ziyue Liu\inst{1, 2}\orcidlink{0009-0004-2793-3326} \and
Federico Girella\inst{1}\orcidlink{0009-0001-6400-8859} \and
Yiming Wang\inst{3}\orcidlink{0000-0002-5932-4371} \and
Davide Talon\inst{3}\orcidlink{0009-0003-6029-1532}
}

\authorrunning{Ziyue Liu, et al.}

\institute{University of Verona, Verona, Italy 
\and Polytechnic University of Turin, Turin, Italy
\and Fondazione Bruno Kessler, Trento, Italy \\
\email{\{ziyue.liu, federico.girella\}@univr.it}\\
\email{\{ywang, dtalon\}@fbk.eu}\\
\vspace{5pt}
\url{https://intelligolabs.github.io/L-VQAScore}
}

\maketitle

\begin{abstract}
Despite the rapid advances in Text-to-Image (T2I) generation models, their evaluation remains challenging in domains like fashion, involving complex compositional generation.
Recent automated T2I evaluation methods leverage pre-trained vision-language models to measure cross-modal alignment. However, our preliminary study reveals that they are still limited in assessing rich entity-attribute semantics, facing challenges in attribute confusion, \ie, when attributes are correctly depicted but associated with the wrong entities. To address this, we build on a Visual Question Answering (VQA) localization strategy targeting one single entity at a time across both visual and textual modalities. We propose a localized human evaluation protocol and introduce a novel automatic metric, \metricfull{} (\metricshort{}), that combines visual localization with VQA probing both correct (reflection) and mislocalized (leakage) attribute generation. 
On a newly curated dataset featuring challenging compositional alignment scenarios, \metricshort{} outperforms state-of-the-art T2I evaluation methods in terms of correlation with human judgments, demonstrating its strength in capturing fine-grained entity-attribute associations. We believe \metricshort{} can be a reliable and scalable alternative to subjective evaluations.
\end{abstract}

\section{Introduction}
Research on Text-to-Image (T2I) generation is rapidly progressing. Recent models~\cite{rombach2022high, podellsdxl} can produce highly detailed and semantically rich images conditioned on natural-language prompts. 
Meanwhile, T2I evaluation is actively evolving, primarily focusing on measuring \textit{the alignment between the conditioning text and the generated image}~\cite{lin2024evaluating, hessel2021clipscore, kirstain2023pick,wu2023human, xu2023imagereward, huang2023t2i, yarom2023you, cho2024davidsonian}. 

\begin{figure}[t] 
    \centering
\includegraphics[width=1\textwidth]{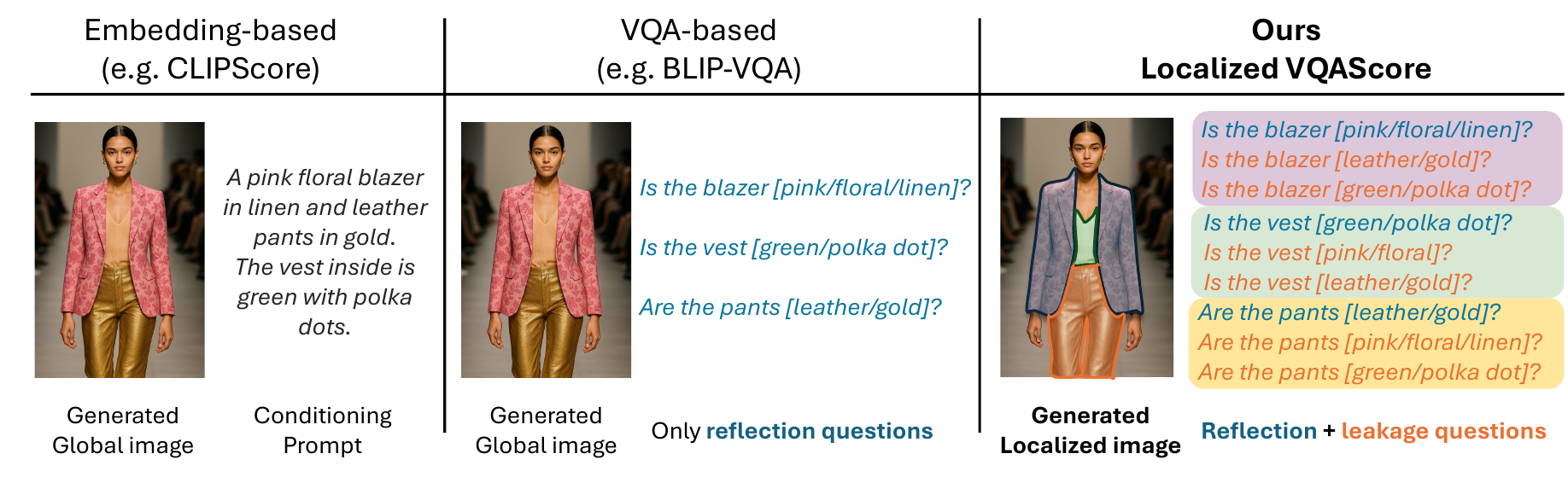}
\vspace{-20pt}
\caption{We investigate Text-to-Image (T2I) evaluation in generated images, particularly in cases concerning compositional prompts, such as in fashion. Existing embedding-based evaluation methods (\eg CLIPScore~\cite{hessel2021clipscore}) struggle to evaluate the entity-attribute bindings correctly. More recent VQA-based methods (\eg BLIP-VQA~\cite{huang2023t2i} or VQAScore~\cite{lin2024evaluating}) improve the compositional understanding by probing whether attributes are reflected in their corresponding entities, with the global image as visual input. Yet, only asking \textit{reflection questions} overlooks the critical \textit{attribute confusion} problem, \ie, when attributes are generated on incorrect entities. Instead, we propose \metricfull{} that leverages localized images and accounts for attribute confusion by probing not only reflection questions, but also \textit{leakage questions} to verify if an attribute is leaked to unrelated entities.}\label{fig:teaser}
\end{figure}

Commonly-adopted \textit{automatic} metrics leverage pre-trained Vision-Language Models (VLMs) by either directly measuring their cross-modal embedding similarity, \eg, CLIPScore~\cite{hessel2021clipscore}, or via Visual Question Answering~(VQA)~\cite{huang2023t2i, yarom2023you, cho2024davidsonian,lin2024evaluating}, see~\cref{fig:teaser}. 
As revealed in recent work~\cite{yuksekgonul2022and, koishigarina2025clip}, VLMs exhibit behaviors akin to bag-of-words models in cross-modal understanding. Thus, they are limited in evaluating compositional semantics with complex entity-attribute bindings, which can be very critical for T2I in domains like fashion.
Recent VQA-based metrics have enhanced the evaluation of entity-attribute binding by checking whether each attribute is correctly reflected in its corresponding entity~\cite{huang2023t2i, yarom2023you, cho2024davidsonian}. However, as highlighted by our preliminary evaluation, existing T2I metrics struggle to recognize \textit{attribute confusion} cases, in other words, when the attributes are reflected on the wrong entities.

On the other hand, human evaluation remains the gold standard in T2I evaluation~\cite{kirstain2023pick, wu2023human, xu2023imagereward}, despite being costly and non-scalable. Yet, 
\textit{can subjective evaluation reliably assess attribute confusion with complex compositional prompts?} Our preliminary study suggests that the answer greatly depends on the adopted evaluation protocol. When requesting human users to evaluate the alignment of the conditioning text and the generated image following the classic 1-5 Likert scale human ratings as done in \cite{otani2023toward,lin2024evaluating,huang2023t2i}, on average, $\sim40\%$ of the time, users disagree about the degree to which an image aligns with the global description. 
Interestingly, when shifting from global to localized assessments, \ie, asking about the presence of a specific entity-attribute pair, human agreement substantially improves, suggesting that attribute confusion can be more reliably evaluated in a localized setup.

Motivated by our preliminary study, we propose an improved human evaluation protocol and an automatic T2I evaluation method in assessing complex prompts with fine-grained semantics. Particularly, we focus on measuring attribute confusion: when a model generates correct entities/attributes, but they are associated incorrectly.
We base our study on fashion data as it naturally presents a wide range of compositional prompts involving multiple garments (entities), \eg, shirts, trousers, jackets, and attributes, \eg, colors, patterns, styles, making it a representative field for compositional challenges. 

Regarding human evaluation, we propose to frame each entity-attribute binding into a binary ``Yes/No'' \textit{reflection question}, \eg, ``\texttt{Is the blazer floral?}'', probing the presence of an expected attribute being reflected on its corresponding clothing item.
To account for attribute confusion, explicit \textit{leakage questions} are also posed to probe the leakage of the attribute to other unrelated entities, \eg, ``\texttt{Is the blazer gold?}'', as shown in~\cref{fig:teaser} (right-most side).
Following similar motivation on localized assessment, we also propose an automatic localization-aware T2I evaluation method, \metricfull{} (\metricshort{}). 
Our approach centers around localized VQA strategies that explicitly evaluate both \textit{reflection questions} and \textit{leakage questions} on localized visual content using segmentation masks. By restricting to the segmented regions, \metricshort{} can better recognize attribute confusion when answering the posed question.  
We evaluate \metricshort{} in terms of rank correlation with human judgments in comparison with existing metrics, achieving the best human alignment with a significant improvement over the prior state-of-the-art. Extensive ablation studies further justify the design choices in terms of the localization strategies and the VQA models.

In summary, the contribution of this work is four-fold:
\begin{itemize}
    \item We investigate and validate the overlooked attribute confusion problem in T2I evaluation with a carefully designed evaluation data covering both automated metrics and human evaluation.
    \item We demonstrate that visual localization and attribute-centric VQA are effective strategies in addressing attribute confusion evaluation.
    \item We propose a new human evaluation protocol and an automated T2I evaluation method \metricshort{}, leveraging both reflection and leakage questions on localized visual content.
    \item \metricshort{} effectively mitigates the attribute confusion in T2I evaluation, achieving improved correlation with human annotations compared to state-of-the-art metrics.
\end{itemize}

\section{Related Work}

\subsection{Text-to-image Generation}
Recent advances in T2I generation have been largely powered by diffusion models~\cite{ho2021classifier}, which synthesize high-quality visuals by simulating a noise-injection (forward process) followed by iterative denoising (backward process)~\cite{nichol2022glide,rombach2022high,ramesh2022hierarchical}. 
Various conditioning strategies have been introduced to improve control over generation, with the most popular being text-guidance. In particular, GLIDE~\cite{nichol2022glide} introduces classifier-free guidance; Stable Diffusion (SD)~\cite{rombach2022high} further introduces cross-attention-based conditioning and performing diffusion in latent space.
More recently, Diffusion Transformers (DiT)~\cite{esser2024scaling}, a novel architecture based on Vision Transformers~\cite{dosovitskiy2020image}, improves generation quality through effective parameter scaling.
\cite{esser2024scaling} popularizes the modeling of a Rectified Flow training objective, substantially increasing generation quality, speed, stability, and prompt adherence.
In this work, we will use recent pre-trained state-of-the-art text-to-image models~\cite{podellsdxl,esser2024scaling, flux2024, stability2024sd35, vivago2025hidream} to investigate the evaluation of attribute confusion.

\subsection{Text-to-image Evaluation Metrics}
The evaluation of T2I models has been a longstanding challenge due to the complex interplay between semantic fidelity, visual quality, and compositionality. 

\noindent\textbf{Visual quality.} Classical perceptual metrics such as FID~\cite{heusel2017gans} has been extensively used for assessing image realism, but fail to capture semantic alignment with the input text.

\noindent\textbf{Global text-image alignment.} More recent approaches leverage contrastive Vision-Language Models (VLMs), notably CLIPScore~\cite{hessel2021clipscore}, which computes cosine similarity between text and image embeddings. However, such metrics are sensitive to lexical variations and often act as “bag-of-words” encoders, thus fail in capturing nuanced relations and attribute bindings~\cite{yuksekgonul2022and, koishigarina2025clip}.

\noindent\textbf{Compositional text-image alignment.} To address these limitations, proposed metrics investigate human-feedback strategies~\cite{xu2023imagereward, kirstain2023pick, wu2023human}, or leveraging Visual Question Answering (VQA) models for automatized global \cite{ku2024viescore, lin2024evaluating} 
or localized ~\cite{cho2024davidsonian, huang2023t2i, yarom2023you}
scoring. 
However, they fail to assess information leakage, and remain prone to attribute confusion, as the visual backbone still processes the full image.

\noindent\textbf{Human evaluation.}
Human evaluation remains a crucial yet inconsistently implemented component in assessing text-to-image generation models~\cite{otani2023toward,karpinska2021perils}.
Key details such as inter-rater agreement are frequently under-reported, and evaluation criteria are often inconsistent and subjective, raising concerns about the reliability and reproducibility of evaluations.
Most implemented user studies focus on overall visual quality~\cite{rombach2022high,nichol2022glide,ding2022cogview2} and global text relevance~\cite{nichol2022glide,ding2022cogview2}, while rarely any assess localization or compositional correctness, overlooking finer-grained issues such as attribute confusion~\cite{otani2023toward}. These emphasize the need for a more standardized and consistent protocol to capture both global quality and fine-grained localization in human evaluation practices.

\section{On T2I Evaluation of Attribute Confusion}\label{sec:preliminary}
This section first formally defines the attribute confusion problem and introduces the data used in our T2I evaluation analysis. Then, we present preliminary studies showing that: (i) localization enables less subjective evaluation when complex attributes are present in the T2I evaluation; and (ii) recent T2I evaluation methods focusing on semantic alignment fail to catch attribute confusion.

\noindent\textbf{Attribute confusion.}
We formally define attribute confusion, a critical problem limiting accurate semantic visual-text alignment~\cite{chefer2023attend,feng2022training,ramesh2022hierarchical}.
Let $P$ be the textual prompt expressed in natural language and let $S$ denote its structured version, $S = \{(e_i, A_i): i=1, .., N\}$ a set of $N\geq0$ entities with each entity $e_i$ associated to a set $A_i$ of $K_i$ attributes.
We refer to attribute confusion when an attribute $a \in A_i$ is associated to a different entity $e_j, j\neq i$ in the generated image.
Attribute confusion occurs 
when a visuo-textual model misassigns attributes to irrelevant regions within an image, resulting in semantically inaccurate results. For instance, when $P$ is expressed as ``\texttt{a pink blazer and gold pants}'', attribute confusion occurs when the T2I model generates the image with pink pants instead of gold pants. Note that while the attribute confusion problem impacts T2I generative models, its automated evaluation requires metrics effectively recognizing correct entity-attribute associations.

\noindent\textbf{Evaluation data.} To conduct human studies and investigate the effectiveness of T2I metrics on attribute confusion, we construct our evaluation data based on Fashionpedia~\cite{jia2020fashionpedia}, a multimodal dataset with fashion images paired with structured item-attribute annotations, suitable for our analysis on attribute confusion.
We select images containing at least two large garments to facilitate easy visual inspection. Recognizing specific fashion attributes, \eg, ``\texttt{notched lapel}'', ``\texttt{set-in}'', might not be straightforward for non-expert users. Thus, we ensure that for each large garment, at least one pattern attribute that is easily recognizable, \eg ``\texttt{striped}'' or ``\texttt{dotted}'', is included in the attribute list. Moreover, we ensure each pattern attribute appears only on a single clothing item to enable easy identification of attribute confusion.
For each image, we then concatenate the garments' attributes with their class labels (\eg, ``\texttt{a striped, notched lapel, long-sleeve shirt}'') and build a conditioning prompt by appending all garment descriptions (\eg, ``\texttt{a striped [\dots] shirt. a pair of dotted [\dots] pants}''). 
With the textual description as conditioning prompt, we then generate images using $5$ state-of-the-art T2I models, namely \texttt{FLUX.1-dev}~\cite{flux2024}, \texttt{SD-3-medium}~\cite{esser2024scaling}, \texttt{SD-3.5-large}~\cite{stability2024sd35}, \texttt{SDXL}~\cite{podellsdxl}, and 4-bit quantized \texttt{HiDream-I1-Full}~\cite{vivago2025hidream}.
Overall, our evaluation data contains 50 outfit descriptions, each featuring at least two entities with unique patterns, and 250 generated images ($5$ per description).

\subsection{Localized Assessment Improves Agreement in Human Evaluation}
As our human study serves as the reference in T2I evaluation, we first investigate how an existing human evaluation protocol handles attribute confusion. We then explore effective strategies exploiting visual localization to improve human evaluation, serving as important inspiration for devising an automated T2I evaluation method. 

\noindent\textbf{Baseline Likert human study.} We first conduct a baseline user study following Likert 1-5 evaluation, a representative protocol widely adopted in T2I evaluation~\cite{otani2023toward}. Specifically, a user evaluates how closely the presented global image aligns with the complete conditioning prompt, on a scale from 1 (weak alignment) to 5 (strong alignment).

\noindent\textbf{Localized human study.} Following the intuition that attending to localized regions could encourage accurate attribute-level assessment, we devise a new human study protocol requiring users to focus on specific entities in the generated image. For an entity, we probe the users with both \textit{reflection questions} and \textit{leakage questions}, where \textit{reflection questions} verify if an attribute is correctly depicted on the entity, while \textit{leakage questions} check if attributes belonging to other entities are miss-localized on the investigated entity, explicitly indicating occurrences of attribute confusion (as demonstrated in~\cref{fig:teaser} (rightmost)). 

\noindent\textbf{Result discussion.} 
On our evaluation data, we collected 1042 and 833 Question-Answer pairs for the baseline Likert and the proposed localized human study, respectively. Then, we measure the agreement among users. Specifically, for each image-question pair, we identify the majority choice among annotators and define user agreement as the average ratio of annotators selecting the majority option.

\begin{wraptable}{r}{0.4\textwidth}
\vspace{-17pt}
\centering
\resizebox{0.95\linewidth}{!}{%
\begin{tabular}{l c}
    \toprule
    \textbf{User Study} & \textbf{Agreement} ($\uparrow$)\\
    \midrule
    Likert~\cite{otani2023toward, huang2023t2i, lin2024evaluating} & 63.5 \\
    \textbf{Localized (Ours)} & \textbf{93.2} \\
    \bottomrule
    \toprule
    \textbf{Method} & \textbf{Failure} ($\downarrow$) \\
    \midrule
    CLIPScore~\cite{hessel2021clipscore} & 46.1 \\ 
    PickScore~\cite{kirstain2023pick} & 29.2  \\ 
    HPSv2Score~\cite{wu2023human} & 29.2 \\
    ImageReward~\cite{xu2023imagereward} & 6.15  \\ 
    BLIP-VQA~\cite{huang2023t2i} & 4.62 \\
    VQAScore\cite{lin2024evaluating} & 4.62  \\
    \textbf{\metricshort{}} \textbf{(Ours)} & \textbf{0.00}  \\
    \bottomrule
\end{tabular}
}
\caption{Pilot study on current evaluation. \textbf{Top:} Agreement rates for user human evaluation studies. \textbf{Bottom:} Failure rate of current T2I evaluation metrics, measured as the percentage of test cases where attribute-swapped pairs receive higher scores.}
\label{tab:preliminary}
\vspace{-27pt}
\end{wraptable}

As shown in~\cref{tab:preliminary} (Top), the classic Likert protocol yields lower agreement, indicating that assessing the alignment between global description and generated image can be challenging and subjective for human evaluators. Complex compositional prompts with multiple attributes often introduce confusion. Moreover, each evaluator has individual preferences and interpretations of the rating criteria, resulting in high variations. 
Differently, the new localized user study achieves a higher agreement, reaching approximately 93\%, a 30\% improvement with respect to Likert. \textit{Localized assessment helps in reducing the complexity and subjectivity in user answering}. Attending to one specific entity at a time allows users to provide more accurate and consistent evaluation on whether an attribute is reflected or leaked.

\subsection{Existing T2I Metrics Fail on Attribute Confusion}
We further examine how state-of-the-art T2I evaluation metrics handle attribute confusion.
We consider both embedding-based metrics (CLIPScore~\cite{hessel2021clipscore}, PickScore~\cite{kirstain2023pick}, HPSv2Score~\cite{wu2023human} and ImageReward~\cite{xu2023imagereward}) and VQA–based metrics (VQAScore~\cite{lin2024evaluating} and BLIP-VQA~\cite{huang2023t2i}), as detailed in~\cref{sec:exp:comparison}.
We conduct a controlled attribute-swapping test. Specifically, for each image-description pair, we swap the attributes belonging to different entities in the description, forming a negative description with swapped attributes, \eg, ``\texttt{a dotted dress and a striped shirt}'' becomes ``\texttt{a striped dress and a dotted shirt}''.
The negative description maintains the same entities and attributes, but with incorrect entity-attribute associations.

We evaluate the metrics of all compared methods using both the correct description and the negative description. We define a metric failure to address attribute confusion when it yields higher alignment scores to images paired with negative descriptions than the correct descriptions. We report the failure rate in~\cref{tab:preliminary} (Bottom).
As expected, embedding-based metrics such as CLIPScore~\cite{hessel2021clipscore}, PickScore~\cite{kirstain2023pick} and HPSv2Score~\cite{wu2023human} inherit the bag-of-words problem from the underlying CLIP~\cite{radford2021learning} backbone, yielding a significantly high failure rate of 46.1\%, 29.2\%, and 29.2\%, respectively.
VQA-based metrics, while mitigating attribute confusion, still present a noticeable ratio of attribute confusion failures. 
By adopting a localized VQA strategy, inspired by our localized human evaluation, the proposed \metricshort{} can avoid failures due to attribute confusion completely on our evaluation data. We describe the metric in detail in the following section.
\section{\metricfull{}}
\begin{figure}[t!]
    \centering
    \includegraphics[width=0.9\linewidth]{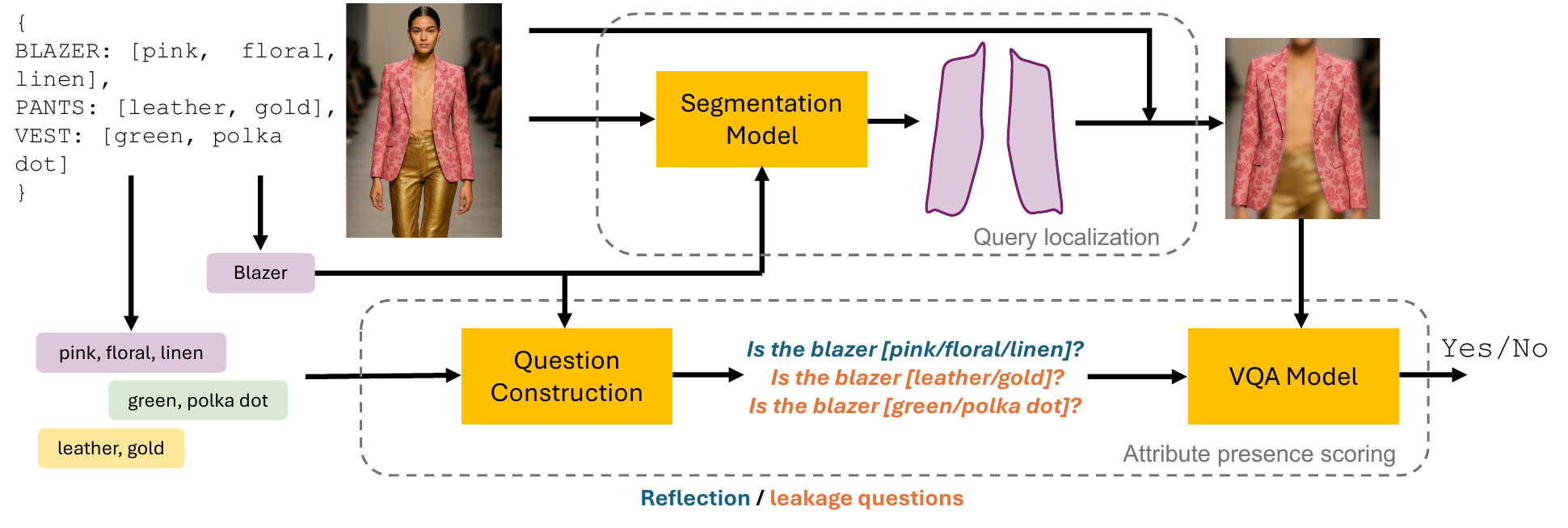}
    \caption{The pipeline of the proposed \metricshort{} in measuring the alignment between the conditioning prompt and the generated image. We represent the conditioning text into structured entity-attribute pairs. 
    \metricshort{} localizes regions of interest leveraging entity categories via a semantic segmentation module. Then reflection and leakage questions are composed to evaluate the presence of desired and leaked attributes in the localized regions, accounting for both attribute depiction  and localization.}
    \label{fig:metric}
\end{figure}

We present \metricfull{}, a novel VQA-based T2I evaluation metric, evaluating attribute reflection and confusion on localized visual content. \metricshort{} scoring can be summarized in three key steps: (i) \textit{localizing the queries} via automatic segmentation, blurring, and cropping of the area of interest, effectively enhancing localization focus (\cref{sec:localizing-queries}); (ii) \textit{scoring the presence of attributes} by querying a state-of-the-art VQA model with both \textit{reflection} and \textit{leakage} questions (\cref{sec:scoring-presence}); and (iii) \textit{metric computation} accounting for both attribute reflection and confusion (\cref{sec:metric-computation}). We refer to~\cref{fig:metric} for an illustration. 

\subsection{Localizing the Queries}
\label{sec:localizing-queries}
To enable more fine-grained evaluation of visual elements and mitigate the problem of attribute confusion evaluation, we introduce a \textit{query localization} approach that explicitly links each attribute to its expected visual region. 
Unlike prior methods that operate over the entire image, our method enforces spatial localization by leveraging entity-level masks obtained from pre-trained segmentation models.

Consider a conditioning textual prompt $P$, describing the desired output image, and its structured version $S$.
For each entity $i, i=1, .., N$, a segmentation mask $M_i \in \{0, 1\}^{H \times W}$ is generated by a pre-trained segmentation model
prompted with the entity class $e_i$.
Formally, a segmentation model $\phi$ segments the generated image $x \in \mathbb{R}^{3 \times H \times W}$: 
\begin{equation}
    M_i = \phi(x, e_i)
\end{equation}
where $M_i$
is the output segmentation mask localizing the class of interest $e_i$ in the visual space.
Then, to minimize the influence of irrelevant visual context on the prediction, we apply a \textit{spatially localized} blurring operation outside mask $M_i$:
\begin{equation}
    \hat{x_i} = M_i \odot x + (1-M_i) \odot \tilde{x}
\end{equation}
where $\odot$ denotes the Hadamard product and $\tilde{x}$ is the processed version of the original image, \ie, its blurred version. 
Finally, we calculate the bounding box $b_i$  containing mask $M_i$ and crop the image around it with some extra margin to accommodate for small segmentation errors. We maintain its original resolution by re-shaping the crop:
\begin{equation}
    x_i = \text{Resize}(\text{Crop}(\hat{x_i}, b_i), H, W)
\end{equation}
where $x_i$ is the final image, $\text{Crop}(\cdot, \cdot)$ is a function that returns image $\hat{x_i}$ cropped around the bounding box $b_i$, and $\text{Resize}(\cdot, \cdot, \cdot)$ is a function that resizes an image to have the longest side matching the given dimension $(H, W)$ and applies white padding to achieve the desired ratio.
\noindent\cref{subsec:ablation} extensively ablates the components of the proposed localization strategy.


\subsection{Scoring the Presence of Attributes}
\label{sec:scoring-presence}
Building on the localization strategy, we then evaluate whether each attribute is both correctly \textit{reflected} and \textit{localized} in the generated image, without attribute confusion. To this end, we employ a Visual Question Answering (VQA) model, and ask visual questions on localized regions.

The set of reflection questions $Q_r$ is constructed from the structured prompt $S$ to verify that attributes explicitly mentioned in the conditioning prompt are correctly reflected within their expected spatial regions:
\begin{equation}
Q_r = \{t(e_i, a, x_i) : \forall a \in A_i, i=1, .., N\},
\end{equation}
where $t(e_i, a, x_i)$ denotes the templated question construction asking about attribute $a$ in entity $e_i$ using the spatially localized visual input $x_i$.

Leakage questions are designed to assess the presence of undesired attributes leaked from other entities, revealing attribute confusion. 
Formally, the set of leakage questions is constructed as: 
\begin{equation}
Q'_l = \{t(e_i, a, x_i) : \forall a \in A_j,  i,j=1, .., N, j \neq i\}.
\end{equation}
We then remove conflicting questions when the same attribute appears for different entities, resulting in $Q_l = Q'_l \setminus Q'_l \cap Q_r$. 

\noindent Finally, building on prior work~\cite{lin2024evaluating}, for all questions $q \in Q = Q_l \cup Q_r$, we evaluate the VQA model \textit{predicted label} in terms of the probability of positive answer to each existence question, given its associated image region as $\textit{Prob}(\text{``Yes''} \mid q).$

\subsection{Metric Computation}
\label{sec:metric-computation}
To quantitatively assess T2I alignment, 
we assign a \textit{target label} to each question based on its semantics, \ie, positive for reflection and negative for leakage questions. 
We then compare the predicted labels to the target ones: a positive reply to a reflection question is counted as a True Positive, whereas a negative answer is counted as a False Negative, \ie, the attribute should have been generated, but it was not. 
Similarly, for leakage questions, a negative reply from the VQA model is counted as a True Negative, while a positive one is counted as a False Positive, \ie, the attribute was not supposed to be generated in this location.
The Positive and Negative answers are the presence/absence of the attributes inside the generated image, thus, only dependent on the generative model used to create $x$.
The final metric evaluates classic \textit{Precision, Recall and F1 Score}, with precision reflecting the model’s ability to avoid hallucinated or misattributed features, and recall capturing the model’s ability to faithfully realize and localize intended attributes. The F1 Score offers a balanced assessment that jointly penalizes omissions and incorrect insertions.

\noindent With little abuse of notation, in the following, we generally refer to \metricshort{} as the collection of both Precision, Recall, and F1 Score metrics,
with the specific metric clarified from the context.

\section{Experiments}\label{sec:experiments}
We first compare \metricshort{} against existing T2I evaluation methods by measuring their correlation to subjective evaluation, following the localized human study protocol and the evaluation data introduced in~\cref{sec:preliminary}.
Next, we present ablation studies to analyze the major designs of \metricshort{}.
\subsection{Comparative Evaluation}\label{sec:exp:comparison}

\noindent\textbf{Baselines.} We consider embedding-based methods including CLIPScore~\cite{hessel2021clipscore} and its human preference–aligned variants,  PickScore~\cite{kirstain2023pick} and HPSv2Score~\cite{wu2023human}, as well as ImageReward~\cite{xu2023imagereward}.
For VQA–based methods, we consider VQAScore~\cite{lin2024evaluating}, evaluating the global alignment through a single visual question, and BLIP-VQA~\cite{huang2023t2i}, which instead assesses alignment by focusing on individual attributes.

\noindent\textbf{Performance measures.}
We employ rank correlation measures to quantify the agreement between automatic metric rankings and the one from the proposed localized human study. 
The human study images are randomly divided into $25$ groups and ranked based on their group F1 Score(Precision/Recall). Similarly, automated metrics rank groups according to their average image-level scores. Results are averaged over $5$ different random seeds.
We report both Spearman's Rho ($\uparrow$) to capture global ranking patterns and Kendall's Tau~($\uparrow$) to capture the pairwise ranking consistency.

\noindent\textbf{Results.}
\cref{tab:performance} presents the performance of \metricshort{} and state-of-the-art approaches. 
The comparison specifically evaluates the extent to which each method aligns with human study results based on model rankings regarding F1 Score, Precision and Recall.  
Accounting for F1 Score, we observe that, overall,
global metrics such as CLIPScore and HPSv2Score exhibit lower performance.
\begin{wraptable}{r}{0.5\textwidth}
\centering
\resizebox{0.95\linewidth}{!}{%
\begin{tabular}{l c c}
        \toprule
        \textbf{Metric} & \makecell{\textbf{Spearman's} \\ \textbf{Rho} ($\uparrow$)} & \makecell{\textbf{Kendall's} \\ \textbf{Tau} ($\uparrow$)}\\
         \midrule
         \multicolumn{3}{c}{\cellcolor{Gray!16}\textbf{Localized Study F1 Score}}\\
        CLIPScore~\cite{hessel2021clipscore} & .460 & .326 \\
        PickScore~\cite{kirstain2023pick} & .433 & .293 \\
        HPSv2Score~\cite{wu2023human} & .215 & .141 \\
        ImageReward~\cite{xu2023imagereward} & .494 & .349 \\
        VQAScore~\cite{lin2024evaluating} & .704 & .536 \\
        BLIP-VQA~\cite{huang2023t2i} & .636 & .492 \\ 

        \textbf{\metricshort{}} \textbf{(Ours)} & \textbf{.818} & \textbf{.650} \\ 
        \multicolumn{3}{c}{\cellcolor{Gray!16}\textbf{Localized Study Precision}}\\
        VQAScore~\cite{lin2024evaluating} & .658 & .504 \\
        \textbf{\metricshort{}  Precision}  \textbf{(Ours)} & .722 & .567 \\
        \multicolumn{3}{c}
        {\cellcolor{Gray!16}\textbf{Localized Study Recall}}\\
        VQAScore~\cite{lin2024evaluating} & .547 & .413 \\
        \textbf{\metricshort{} Recall}  \textbf{(Ours)} & .768 & .670 \\
        \bottomrule
    
\end{tabular}
}
\caption{Performance in T2I alignment regarding the localized study F1 Score, Precision and Recall.
    \metricshort{} consistently surpasses existing state-of-the-art methods. 
    }\label{tab:performance}
\vspace{-15pt}
\end{wraptable}
Among these embedding-based methods, ImageReward, which incorporates fine-tuning of a regressor on human preference data, demonstrates relatively improved performance.
Notably, \metricshort{} consistently outperforms the strongest VQA-based metrics, VQAScore and BLIP-VQA, by around 16\% and 26\% respectively. 
The results demonstrate the effectiveness of \metricshort{} building on localized assessment to probe attribute confusion. Similarly, the improved correlation is observed with Localized Study Precision/Recall metrics, where VQAScore fails to catch the nuances of reflection and leakage questions.

\subsection{Ablation Study}\label{subsec:ablation}
\noindent\textbf{Localization strategy.} 
We investigate how different localization strategies
affect \metricshort{} correlation with the localized human study: i) without localization, ii) segmentation-based \textit{masking} by blacking out the non-target regions, iii) \textit{blurring} of surrounding context, iv) \textit{cropping} according to the mask bounding box, as well as v) sequentially \textit{Masking and Cropping}, and vi) the proposed \textit{Blurring and Cropping}.
We further experiment with \textit{Blurring and Cropping w/ OV-SEG}, relying on 
\begin{wraptable}{r}{0.55\textwidth}
\vspace{-20pt}
\centering
\resizebox{0.95\linewidth}{!}{%
        \begin{tabular}{lcc}
    \toprule
    \textbf{Localization Strategy} & \makecell{\textbf{Spearman's} \\ \textbf{Rho} ($\uparrow$)} & \makecell{\textbf{Kendall's} \\ \textbf{Tau} ($\uparrow$)}\\
    \midrule
    \textcolor{red}{\normalsize \ding{55}} & .549 & .416\\
    Masking & .697 & .527 \\
    Blurring & .742 & .579 \\
    Cropping & .682 & .519 \\
    Masking and Cropping & .723 & .553 \\
    \makecell[l]{Blurring, Cropping w/ OV-SEG} & .682 & .546 \\ 
    Blurring, Cropping \textbf{(Ours)} & \textbf{.818} & \textbf{.650} \\
    \bottomrule
    \toprule
        \textbf{VQA Model} & \makecell{\textbf{Spearman's} \\ \textbf{Rho} ($\uparrow$)} & \makecell{\textbf{Kendall's} \\ \textbf{Tau} ($\uparrow$)}\\
        \midrule
        LLaVA-v1.5-Vicuna-7b~\cite{liu2024improved} & .660 & .500 \\
        InstructBLIP-Flan-T5-xl~\cite{dai2023instructblip} & .715 & .570 \\
        CLIP-Flan-T5-xxl~\cite{lin2024evaluating} & \textbf{.818} & \textbf{.650}\\
        \bottomrule
    \end{tabular}
    }
    \caption{Ablation analysis on \metricshort{}. \textbf{Top:} the effect of localization strategy.\\ \textbf{Bottom:} the choice of VQA model.}\label{tab:ablation-vlm}
    \vspace{-20pt}
\end{wraptable}
OV-SEG~\cite{liang2023open} for segmentation, as opposed to our adopted Grounded-SAM-2~\cite{ren2024grounded,ravi2024sam}.
We report results in~\cref{tab:ablation-vlm} (Top). Segmentation-based \textit{Blurring} consistently outperforms hard exclusion techniques such as \textit{Masking} and \textit{Cropping}, while our \textit{Blurring and Cropping} yields further improvements. We hypothesize that this combination effectively balances the trade-off between context preservation and spatial focus. 
Crucially, the choice of the segmentation model affects the metric alignment to user evaluation. 
Grounded-SAM-2~\cite{ren2024grounded,ravi2024sam} is superior to OV-SEG in localization accuracy. Qualitative investigation suggests that OV-SEG struggles to segment fashion-related entities.

\noindent\textbf{VQA model.} 
As shown in~\cref{tab:ablation-vlm} (Bottom), we examined the impact of the VQA models.
Consistent with other VQA-based methods~\cite{huang2023t2i,lin2024evaluating}, the underlying VLM model has a notable impact on results. 
Our approach is orthogonal to the choice of VQA model, allowing flexible integration with any backbone and implying that future stronger VQA models can further enhance \metricshort{}.

\section{Conclusions}
In conclusion, this work identifies and addresses a critical gap in text-to-image  evaluation, namely the challenge of attribute confusion. By introducing a localized evaluation framework centered on spatially-aware VQA, the proposed \metricshort{} metric enables precise assessment of whether attributes are correctly realized and localized in generated images. Our findings reveal that current evaluation metrics and human protocols fail to capture fine-grained semantic mismatches, particularly in multi-entity scenarios. Thus, we introduce a novel human evaluation protocol targeted explicitly at the attribute confusion problem. \metricshort{} demonstrates stronger correlation with human judgments, providing a robust, automated solution for detecting mislocalized attributes and offering a meaningful step forward in evaluating compositionality in T2I systems.

\smallskip
\noindent\textbf{Acknowledgment.} This study was supported by LoCa AI, funded by Fondazione CariVerona (Bando Ricerca e Sviluppo 2022/23), PNRR FAIR - Future AI Research (PE00000013) and Italiadomani (PNRR, M4C2, Investimento 3.3), funded by NextGeneration EU.
We acknowledge the CINECA award under the
ISCRA initiative, for the availability of high-performance computing resources and support.
We acknowledge EuroHPC Joint Undertaking for awarding us access to MareNostrum5 as BSC, Spain.
Finally, we acknowledge HUMATICS, a SYS-DAT Group company, for their valuable contribution to this research.

%
%
\bibliographystyle{splncs04}
\bibliography{main}
\end{document}